\documentclass[conference]{IEEEtran}
\IEEEoverridecommandlockouts
\usepackage{cite}
\usepackage{amsmath,amssymb,amsfonts}
\usepackage{algorithmic}
\usepackage{graphicx}
\usepackage{textcomp}
\usepackage{xcolor}
\usepackage{array}
\usepackage{amsmath}
\usepackage{amssymb}
\usepackage{booktabs}
\usepackage{adjustbox} 
\usepackage{booktabs} 
\usepackage{multirow} 
\usepackage[T1]{fontenc}
\usepackage[utf8]{inputenc} 
\DeclareUnicodeCharacter{2032}{$^\prime$}

\usepackage{subfig}
\usepackage{subcaption}
\usepackage{graphicx}
\usepackage[hidelinks]{hyperref}

\hypersetup{
    colorlinks=true,
    linkcolor=blue,
    filecolor=blue,      
    urlcolor=cyan,
    pdftitle={Overleaf Example},
    citecolor=black,
    pdfpagemode=FullScreen,
    }
\def\BibTeX{{\rm B\kern-.05em{\sc i\kern-.025em b}\kern-.08em
    T\kern-.1667em\lower.7ex\hbox{E}\kern-.125emX}}
\begin{document}

\title{Depth Jitter: Seeing through the Depth\\}
\author{\IEEEauthorblockN{Md Sazidur Rahman}
\IEEEauthorblockA{
\textit{Université de Toulon}\\
Toulon, France \\
\textit{Instituto Superior Técnico}\\
Lisbon, Portugal\\
mdsazidur02@gmail.com}
\and
\IEEEauthorblockN{David Cabecinhas}
\IEEEauthorblockA{\textit{Institute for Systems and Robotics} \\
\textit{Instituto Superior Técnico}\\
Lisbon, Portugal \\
dcabecinhas@isr.tecnico.ulisboa.pt}
\and
\IEEEauthorblockN{Ricard Marxer}
\IEEEauthorblockA{
\textit{Université de Toulon,}\\
\textit{ Aix Marseille University,}\\
\textit{CNRS, LIS}\\
Toulon, France \\
marxer@univ-tln.fr}
}

\maketitle

\begin{abstract}
Depth information is essential in computer vision, particularly in underwater imaging, robotics, and autonomous navigation. However, conventional augmentation techniques overlook depth aware transformations, limiting model robustness in real world depth variations. In this paper, we introduce Depth-Jitter, a novel depth-based augmentation technique that simulates natural depth variations to improve generalization. Our approach applies adaptive depth offsetting, guided by depth variance thresholds, to generate synthetic depth perturbations while preserving structural integrity. We evaluate Depth-Jitter on two benchmark datasets, FathomNet and UTDAC2020 demonstrating its impact on model stability under diverse depth conditions. Extensive experiments compare Depth-Jitter against traditional augmentation strategies such as ColorJitter, analyzing performance across varying learning rates, encoders, and loss functions. While Depth-Jitter does not always outperform conventional methods in absolute performance, it consistently enhances model stability and generalization in depth-sensitive environments. These findings highlight the potential of depth-aware augmentation for real-world applications and provide a foundation for further research into depth-based learning strategies. The proposed technique is publicly available to support advancements in depth-aware augmentation. The code is publicly available on \href{https://github.com/mim-team/Depth-Jitter}{github}.
\end{abstract}


\section{Introduction}
\noindent Marine life significantly predates terrestrial life, with evidence suggesting its origins approximately 3.7 billion years ago \cite{dodd_evidence_2017}, while terrestrial life emerged around 3.1 billion years ago \cite{battistuzzi_major_2009}. The fossil record further indicates that marine biodiversity has outpaced terrestrial diversity for most of Earth's history \cite{benton_origins_2016}. Covering 71\% of the planet, oceans serve as vast ecosystems that support immense species richness, aligning with biogeographic theories linking habitat size to biodiversity \cite{costello_surface_2010}. Despite its ecological significance, deep-sea research is hindered by extreme conditions, high costs, and technological limitations. Traditional exploration relies on resource-intensive submersibles and remotely operated vehicles (ROVs), but recent advancements in Autonomous Underwater Vehicles (AUVs) and submarine gliders, equipped with acoustic sensors and high-resolution imaging, have greatly expanded deep-sea mapping capabilities \cite{yoerger_surveying_1998,yoerger_autonomous_2007,henthorn_high-resolution_2006,german_hydrothermal_2008}.  

\noindent The integration of deep learning has further enhanced underwater research, automating species identification, debris classification, and habitat assessment. Convolutional Neural Networks (CNNs) enable AUVs to autonomously analyze marine environments, streamlining biodiversity monitoring and conservation efforts by reducing the need for manual analysis.  

\noindent This study is motivated by the vast, unexplored deep ocean, which harbors rare species and plays a vital role in global ecological processes. Traditional marine observation methods are constrained by human accessibility, environmental factors, and financial constraints. To address these challenges, we propose integrating deep learning with depth-aware image augmentation techniques, enhancing underwater image analysis and classification.  

\noindent Ultimately, this research seeks to advance deep-sea exploration by leveraging AI-driven methodologies. By improving species recognition and underwater scene interpretation, our approach contributes to marine biodiversity research, conservation, and a deeper understanding of Earth's least explored frontier.

\subsection{Contribution}
\noindent This research explores techniques to enhance underwater object classification by leveraging multilabel classification and depth-aware augmentation. The key contributions of this work are as follows:

\begin{itemize}
    \item We propose a novel \textbf{depth-based augmentation technique} tailored for underwater datasets, designed to improve the robustness and accuracy of object classification and detection in varying underwater conditions.
    \item We establish a \textbf{comprehensive benchmark} on two diverse underwater datasets, demonstrating the effectiveness of our proposed augmentation method. This benchmark serves as a foundation for future research in depth-aware learning for underwater computer vision.
    \item We analyze the impact of depth-aware transformations on multilabel classification performance, providing \textbf{valuable insights into augmentation strategies} for deep-sea imaging and marine biodiversity monitoring.
\end{itemize}
\section{Background}
\subsection{Multi-label Classification}
\noindent Multi-label classification assigns more than one label to an image, making it particularly useful for underwater environments where multiple species or objects may coexist within a scene.  

\noindent Early approaches, such as Binary Relevance (BR) \cite{boutell_learning_2004}, treated each label as an independent classification task. While simple, this method ignored label correlations, limiting its effectiveness. Classifier Chains (CC) \cite{read_classifier_2011} addressed this limitation by modeling dependencies between labels, whereas RAkEL \cite{rakel} introduced a randomized ensemble approach to improve label correlation modeling.  

\noindent Recent advancements have integrated deep learning with label dependency structures. The CNN-RNN framework \cite{wang2016cnnrnn} combines convolutional neural networks (CNNs) for feature extraction with recurrent neural networks (RNNs) to capture label dependencies. Feng Zhu et al. \cite{ZhuLOYW17} proposed a Spatial Regularization Network, which models spatial relationships among labels, enhancing classification performance. Renchun You et al. \cite{CMA} introduced a framework utilizing cross-modality attention and semantic graph embeddings to improve multi-label classification, demonstrating state-of-the-art results in complex environments.

\subsection{Underwater Object Detection and Classification}
\noindent Object detection in underwater environments presents unique challenges due to variable lighting, visibility degradation, and the dynamic movement of marine organisms. Traditional techniques relied on manually engineered features \cite{duan_2015}, but these approaches struggled with real-world adaptability due to their dependence on domain-specific heuristics.  

\noindent The advent of deep learning has significantly improved underwater object detection by enabling models to learn data-driven feature representations. Wu et al. \cite{wu2019research} investigated the impact of illumination on underwater detection performance, revealing that light variations significantly affect model accuracy. Peng et al. \cite{peng2021review} reviewed deep learning-based preprocessing techniques, highlighting that while they enhance visibility, architectural advancements and domain-specific integration remain key areas for improvement. Recently, Jian et al. \cite{jian_underwater_2024} conducted a comprehensive survey on object detection and tracking techniques in underwater imaging, providing a comparative analysis of existing methodologies.  

\subsection{Improving Model Robustness through Data Augmentation}
\noindent Data augmentation plays a crucial role in enhancing model robustness, particularly in challenging underwater environments where data collection is expensive and logistically complex. Augmentation techniques artificially expand datasets, improving generalization to unseen conditions.  

\noindent Geometric transformations such as rotation, scaling, and flipping help models learn invariance to different perspectives. Color transformations (e.g., brightness, contrast, hue adjustments) simulate underwater lighting variations, a key challenge in marine imaging. Cutout techniques, where random portions of an image are obscured, improve robustness to occlusion caused by marine organisms or vegetation.  

\noindent Additionally, Generative Adversarial Networks (GANs) have been explored for generating synthetic underwater imagery, augmenting training datasets where real-world data is scarce. Such approaches, when integrated with traditional augmentations, significantly enhance the robustness, adaptability, and classification performance of models in complex underwater conditions.

\section{Methodology}

\noindent Underwater imagery poses challenges due to light absorption and scattering, leading to inconsistent colors and degraded object visibility. This variability complicates object classification in marine environments.To enhance model robustness, we introduce a depth-aware augmentation technique based on underwater image formation models. Our method estimates scene depth and extracts key imaging parameters, applying depth offsets to re-render images with realistic underwater variability. This augmentation enriches the training set, improving species classification by increasing resilience to domain shifts.
\noindent Additionally, we implement stratified sampling to address class imbalance in multi-label classification, ensuring better representation of underrepresented species and reducing bias toward dominant classes.

\subsection{Underwater Image Formation Model}

\noindent Underwater imaging is affected by \textit{light absorption} and \textit{scattering}, which distort colors based on object distance and water properties. To correct these distortions, we utilize the \textbf{Underwater Image Formation Model (UIFM)} \cite{boittiaux:tel-04482249}, which describes how light interacts with the medium.

\noindent The observed underwater image intensity is given by:

\begin{equation}
\label{uifm_eq}
I_{c,p} = J_{c,p}e^{-\beta_c z_p} + B_c(1 - e^{-\gamma_c z_p}),
\end{equation}

\noindent where:
\begin{itemize}
    \item \( I_{c,p} \) and \( J_{c,p} \) are the observed and restored intensities for color channel \( c \).
    \item \( B_c \) represents the veiling light (background light).
    \item \( \beta_c \) and \( \gamma_c \) denote the \textit{absorption} and \textit{backscatter coefficients}.
    \item \( z_p \) represents the scene depth at pixel \( p \).
\end{itemize}

\noindent Akkayanak et al. \cite{seathru} refined this model to distinguish between absorption and backscatter, improving color restoration. \textit{Gaussian SeaThru} \cite{boittiaux:tel-04482249} further optimized parameter estimation by introducing statistical priors.

\noindent The restored image is computed by solving for \( J_{c,p} \):

\begin{equation}
J_{c,p} = \left( I_{c,p} - B_c (1 - e^{-\gamma_c z_p}) \right) e^{\beta_c z_p}.
\end{equation}

\noindent These parameters (\( B_c, \beta_c, \gamma_c \)) are estimated by minimizing a likelihood function derived from image statistics, ensuring a more accurate restoration of underwater colors.

\begin{figure}[htbp]
  \centering
  \subfloat[Original image]{\includegraphics[width=0.3\linewidth]{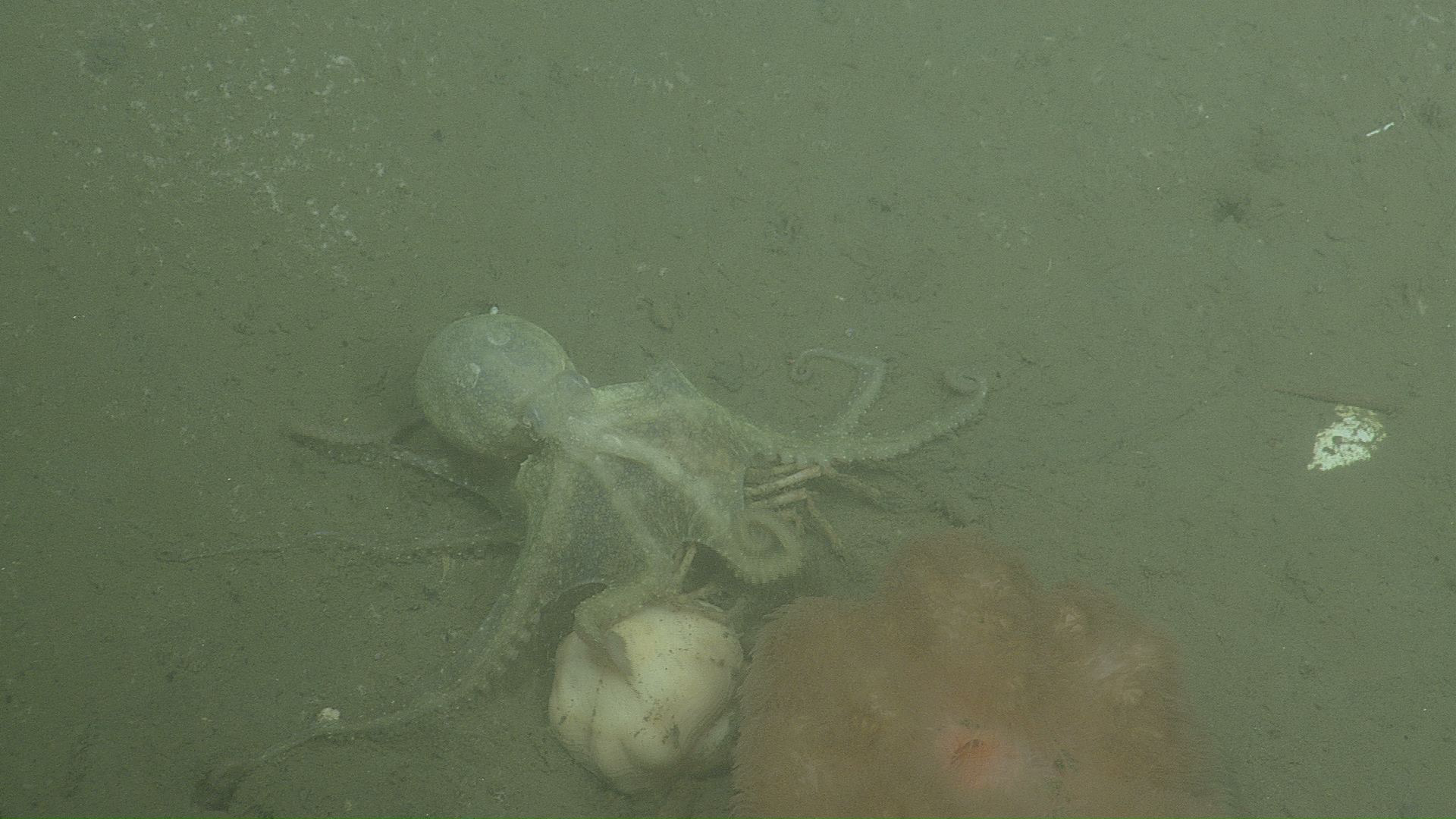}} \hfill
  \subfloat[Depth map]{\includegraphics[width=0.3\linewidth]{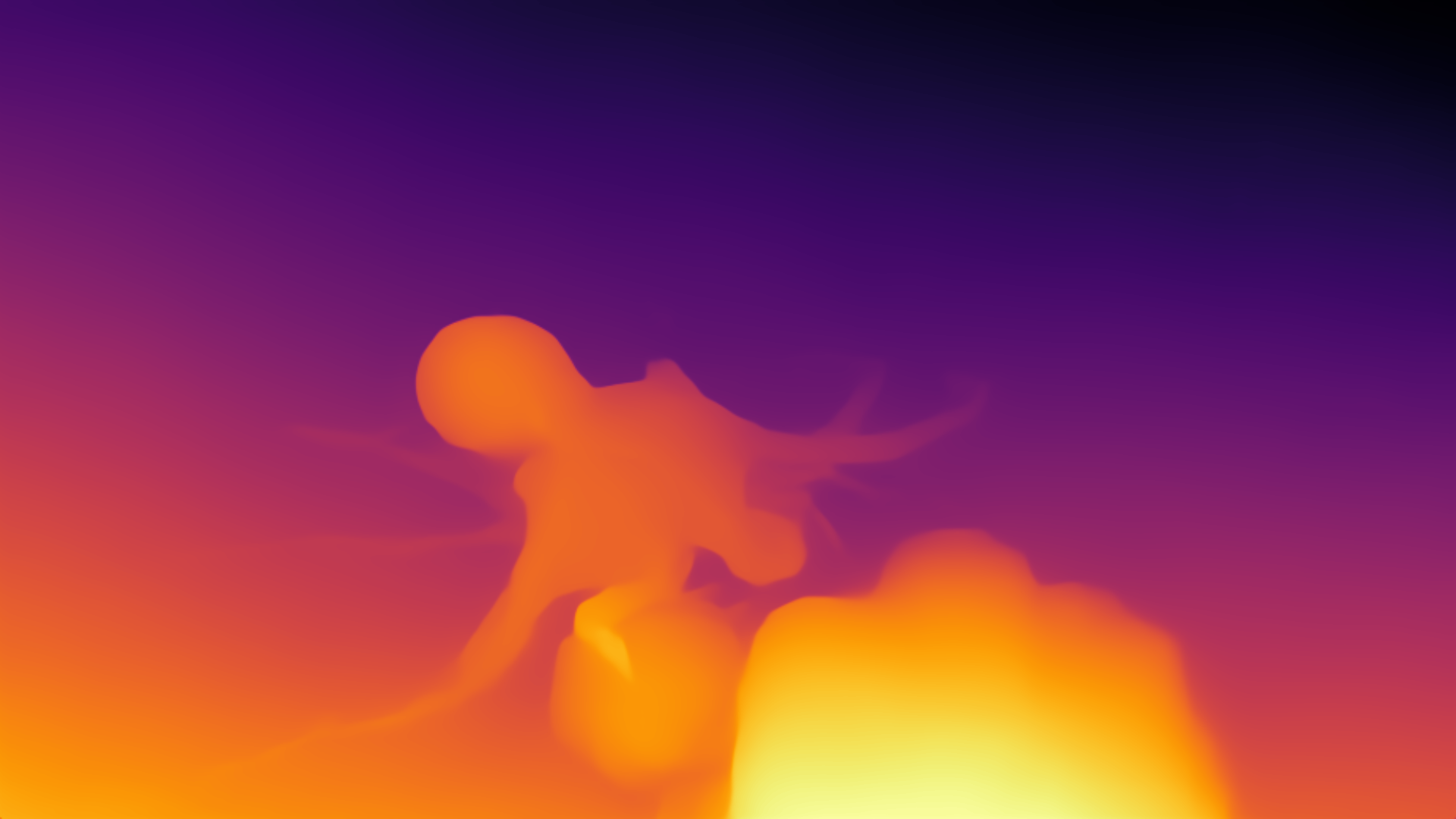}} \hfill
  \subfloat[Restored image]{\includegraphics[width=0.3\linewidth]{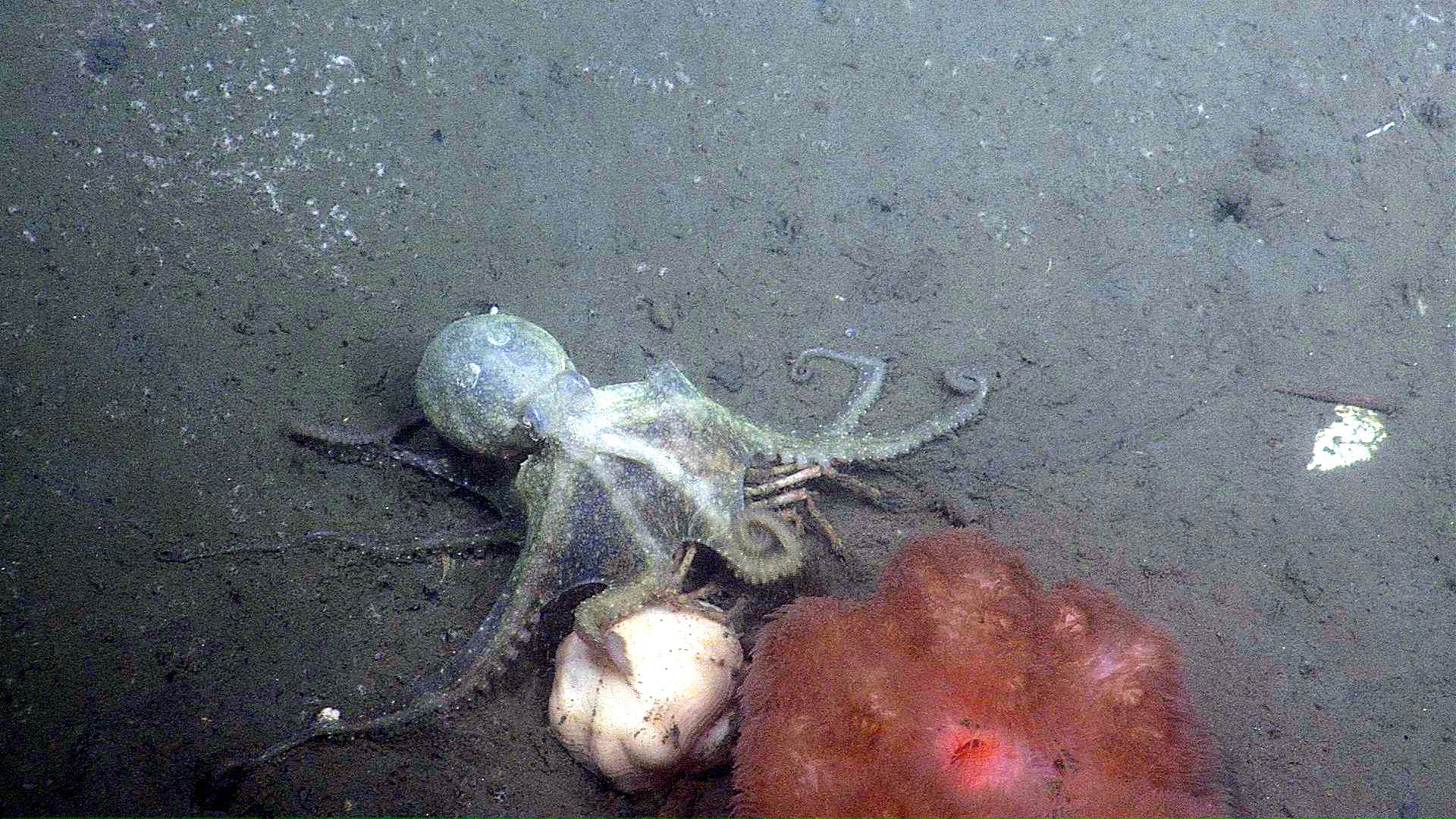}} \\
  \subfloat[Original image]{\includegraphics[width=0.3\linewidth]{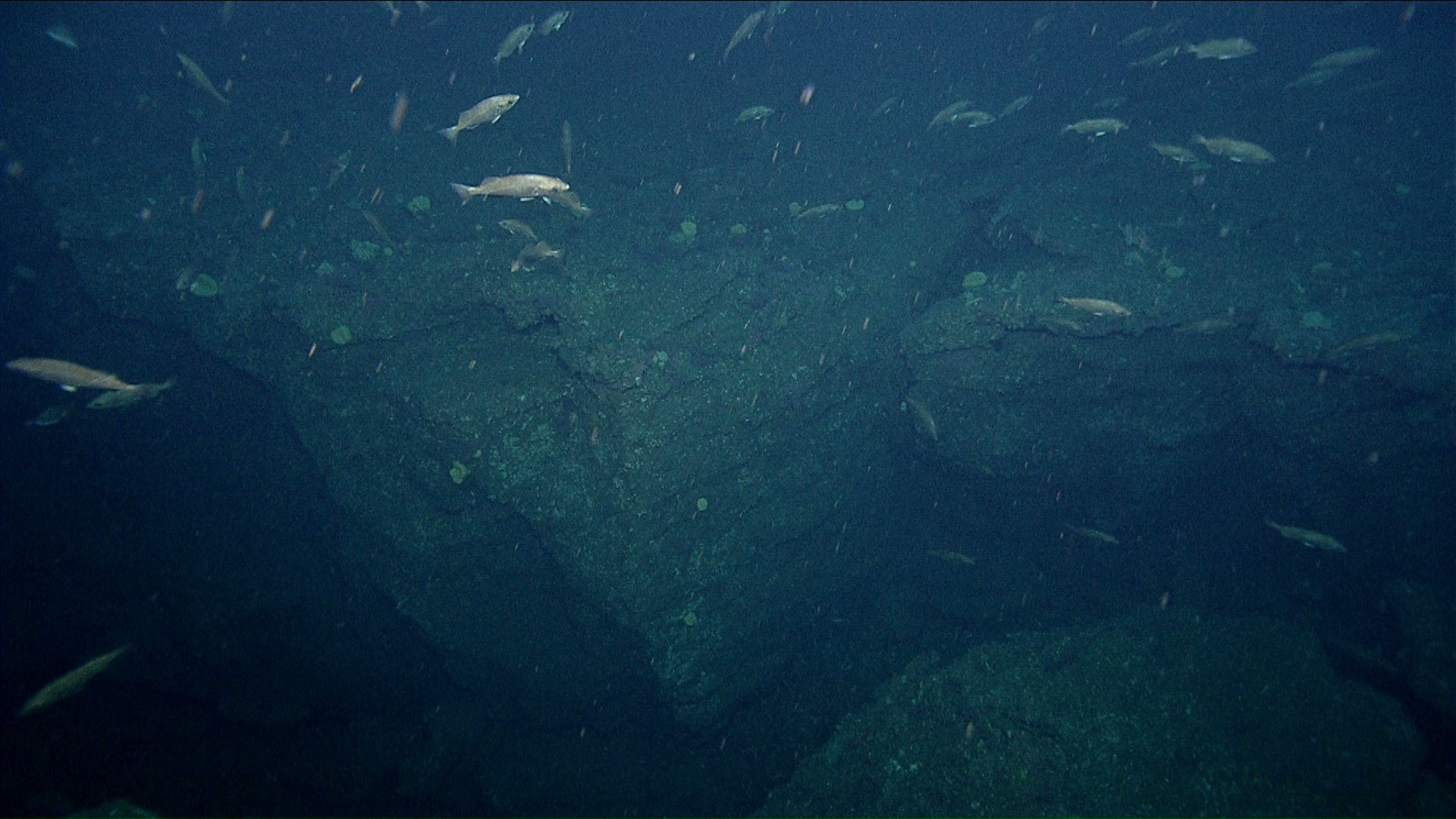}} \hfill
  \subfloat[Depth map]{\includegraphics[width=0.3\linewidth]{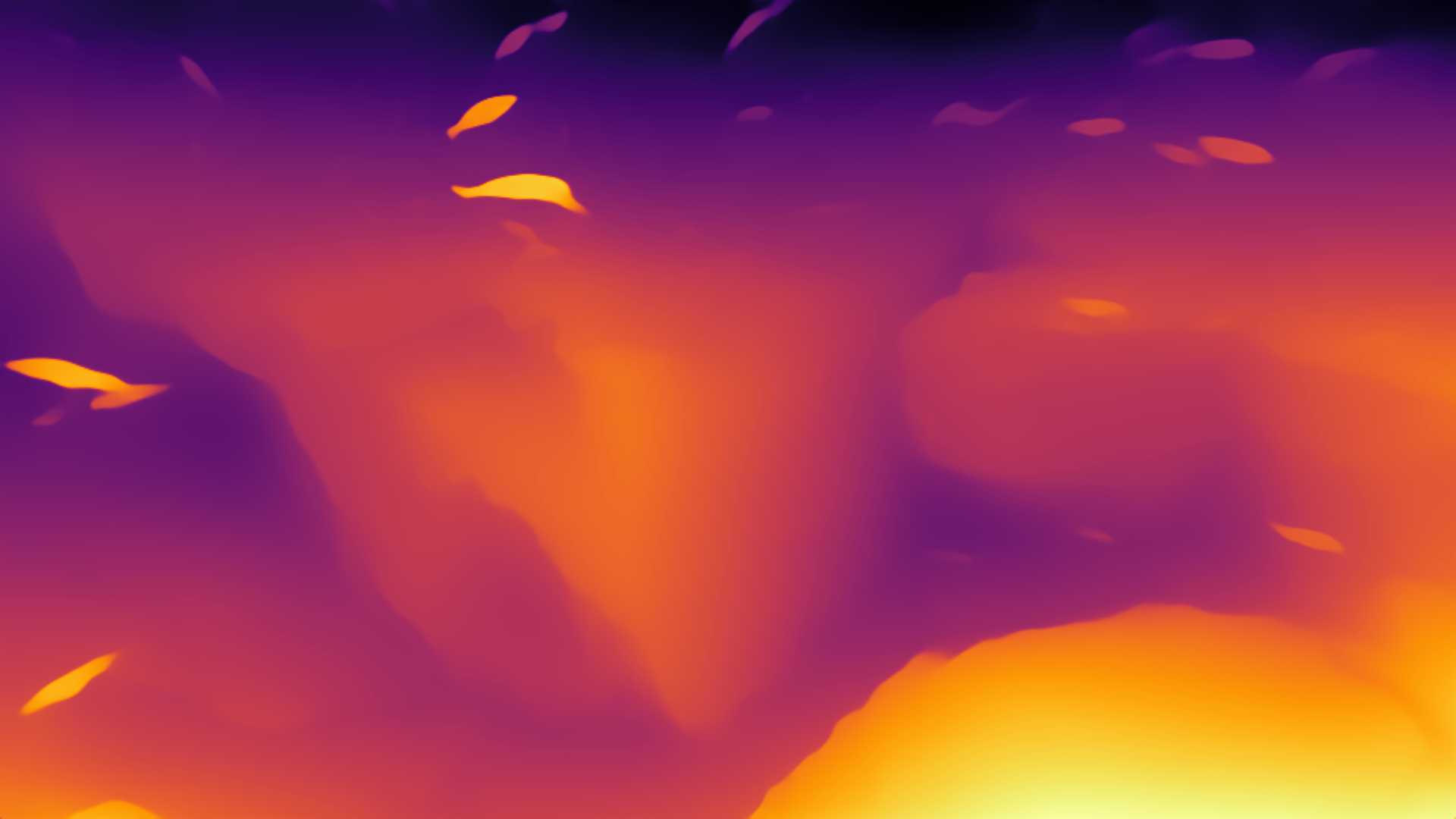}} \hfill
  \subfloat[Restored image]{\includegraphics[width=0.3\linewidth]{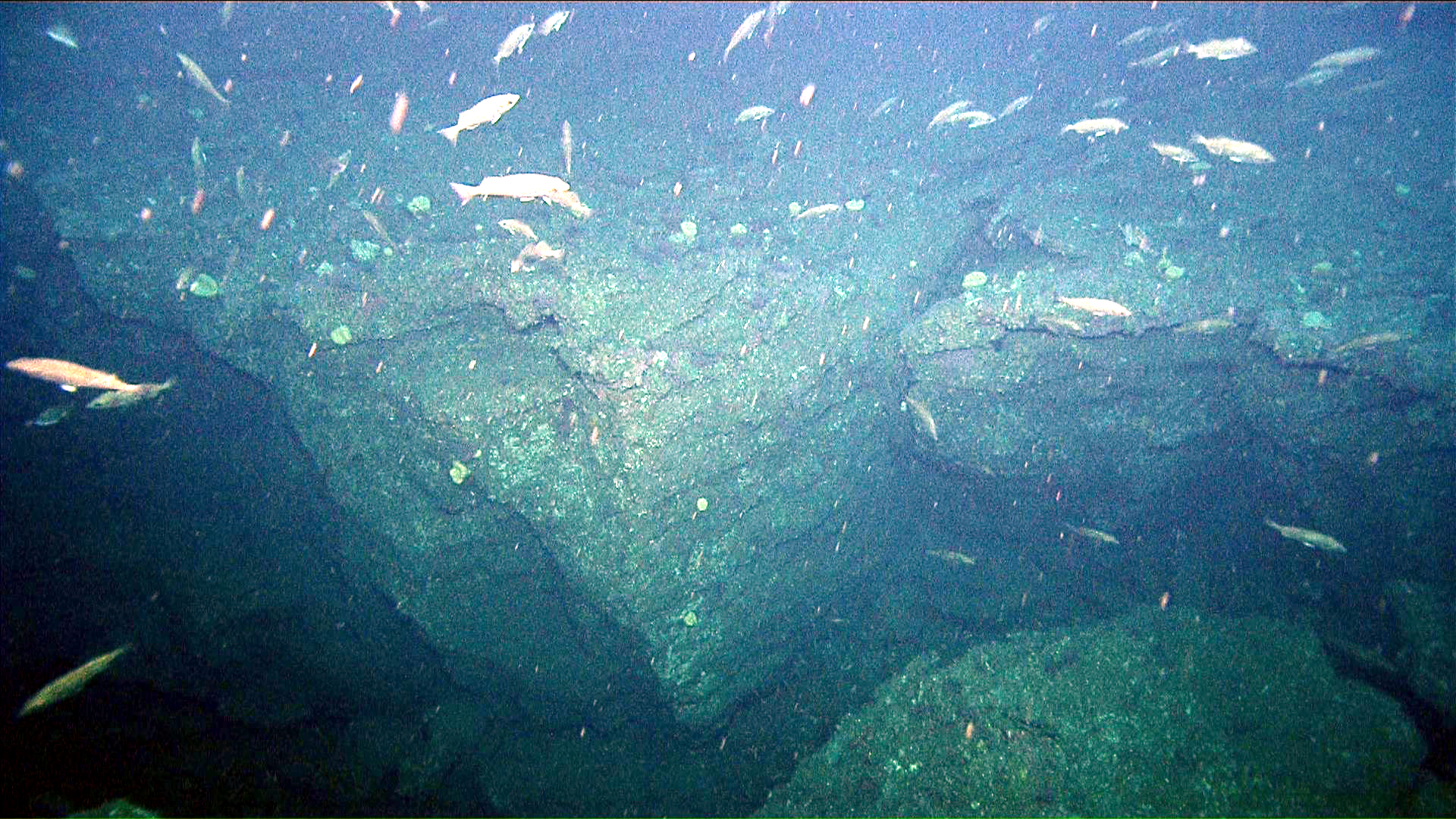}}
  \caption{Examples of restored images using UIFM: original image (left), estimated depth map (middle), and restored image (right).}
  \label{fig:fig1}
\end{figure}

\noindent By leveraging \textbf{Depth-Anything} \cite{depthAnything} for depth estimation and \textit{SeaThru}-based restoration, we obtain depth-aware image transformations that enrich training data diversity. This serves as the foundation for our \textbf{Depth Jitter} augmentation, where controlled depth offsets generate synthetic variations in scene depth.

\subsection{Depth Jitter}

\noindent To enhance multi-label classification performance, we employ the underwater image formation model described in Equation \ref{uifm_eq} as an on-the-fly augmentation technique during training. This method dynamically modifies pixel intensities based on depth variations, thereby improving model robustness to environmental changes in underwater imagery.  

\noindent We optimize the computation of $I_{c,p}$ by introducing a depth offset into the image formation model. Let $I_{c,p}^{orig}$ denote the original image and $I_{c,p}^{mod}$ the depth-modified image. Using the underwater image formation model, they are expressed as:
\begin{equation}
\label{I_orig}
I_{c,p}^{orig} = J_{c,p}e^{-\beta_c z_p} + B_c(1 - e^{-\gamma_c z_p}),
\end{equation}
\begin{equation}
\label{I_mod}
I_{c,p}^{mod} = J_{c_p}e^{-\beta_c z_m} + B_c(1 - e^{-\gamma_c z_m})
\end{equation}
By solving for $J_{c,p}$ in  \eqref{I_orig}:
\begin{equation}
J_{c,p} = \frac{I_{c,p}^{orig} - B_c (1 - e^{-\gamma_c z_p})}{e^{-\beta_c z_p}},
\end{equation}
and substituting it into \eqref{I_mod}, we obtain:
\begin{equation}
I_{c_p}^{mod} = \left( \frac{I_{c,p}^{orig} - B_c (1 - e^{-\gamma_c z_p})}{e^{-\beta_c z_p}} \right) e^{-\beta_c z_m} + B_c (1 - e^{-\gamma_c z_m})
\end{equation}
This simplifies to:
\begin{equation}
\label{I_mod_final}
I_{c_p}^{mod} = \left( I_{c_p}^{orig} - B_c (1 - e^{-\gamma_c z_p}) \right) e^{-\beta_c (\Delta z_m - z_p)} + B_c (1 - e^{-\gamma_c \Delta z_m})
\end{equation}

\noindent Here, $\Delta z_m$ represents the depth offset applied to the original depth map. By introducing these offsets, we generate synthetic training samples with varying depths, effectively simulating different underwater conditions. This augmentation enhances the model’s ability to generalize across diverse lighting and turbidity levels, reducing domain shift issues in real-world deployment.  

\begin{figure}[htbp]
    \centering
    \includegraphics[width=0.5\textwidth]{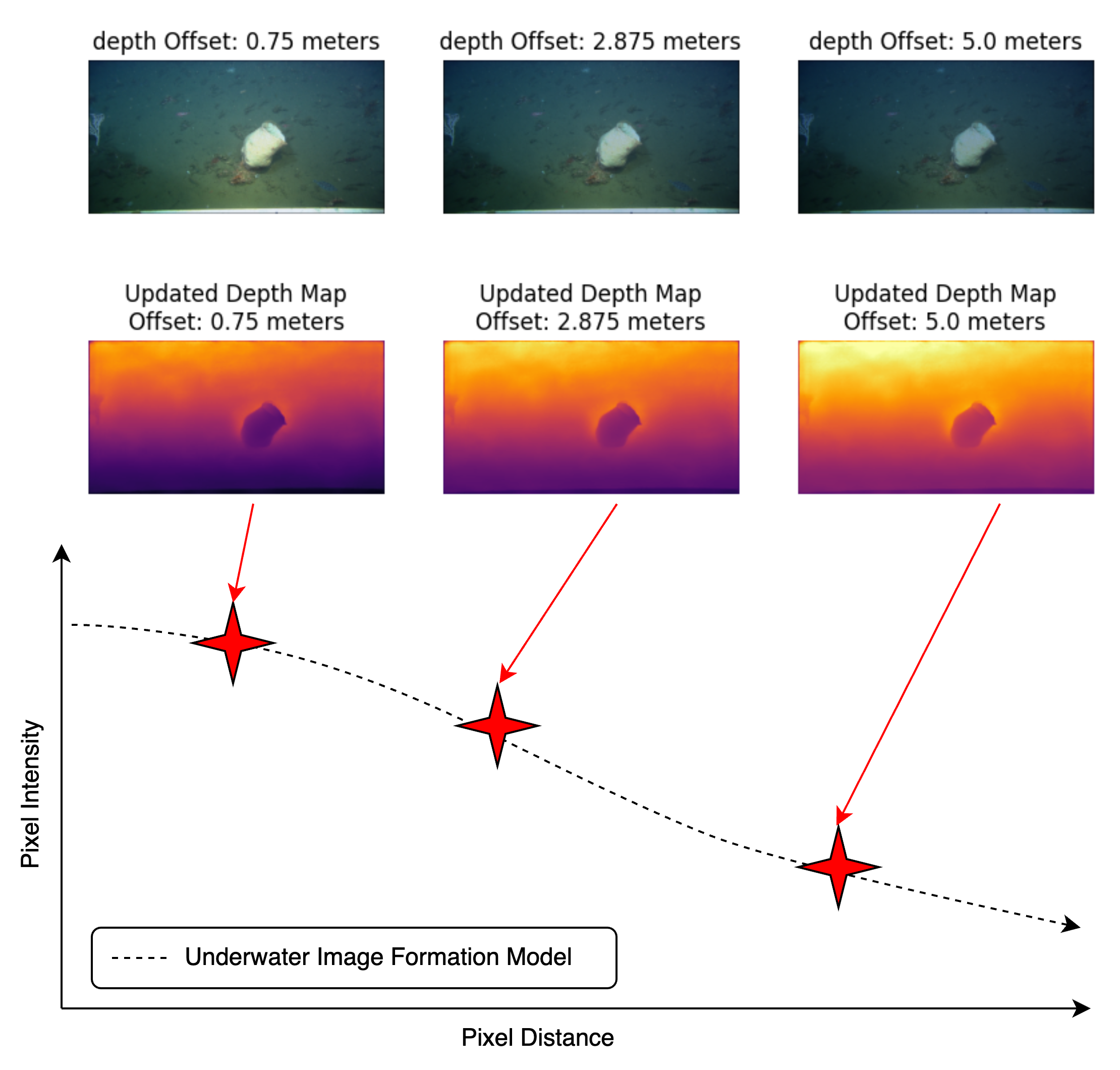}
    \caption{Pixel intensity variations under different depth settings.}
    \label{fig:depthJitter}
\end{figure}

\noindent Figure \ref{fig:depthJitter} illustrates how pixel intensity varies with depth. Greater depths result in lower intensity values, leading to a darker image, whereas shallower depths increase pixel intensity and color vibrancy. These intensity variations are evident in the corresponding depth maps, demonstrating the impact of depth-based augmentation on image formation.

\noindent It is important to note that parameter estimation in the Underwater Image Formation Model (UIFM) and DepthAnything inference can be computationally intensive. However, since these estimates are computed during the pre-processing stage, the only operation performed during training is the re-rendering of images using adjusted depth offsets in Equation \ref{I_mod_final}. This significantly reduces computational overhead, making our Depth Jitter augmentation strategy both effective and efficient during training.

\subsection{Offsetting Strategy for Depth Augmentation}
\noindent Depth variability significantly impacts the effectiveness of our augmentation approach. To handle this variability systematically, we employ a quantile-based thresholding technique, defining the depth variance threshold \( \tau \) as the 25th percentile of the variance distribution:

\begin{equation}
    \tau = Q_{0.25}(V),
\end{equation}

 \noindent where \( Q_{0.25}(V) \) represents the first quartile of the empirical depth variance distribution \( V \). 


\noindent Given an image \( I \) with depth variance \( v \), we categorize it as:

\begin{equation}
    I \in 
    \begin{cases} 
      \text{High-variance set}, & \text{if } v \geq \tau, \\
      \text{Low-variance set}, & \text{otherwise}.
    \end{cases}
\end{equation}

\noindent Since low-variance images converge towards depth boundaries, Depth Jitter does not significantly alter their characteristics. Thus, augmentation is applied only to high-variance images:

\begin{equation}
    \Delta z_m \sim 
    \begin{cases} 
      \mathcal{U}(-\alpha z_{\min}, \beta z_{\max}), & \text{if } v \geq \tau, \\
      0, & \text{if } v < \tau.
    \end{cases}
\end{equation}

\noindent where:
\begin{itemize}
    \item \( \Delta z_m \) is the randomly applied depth offset.
    \item \( z_{\min} \) and \( z_{\max} \) denote the minimum and maximum depths in the image.
    \item \( \mathcal{U}(a, b) \) denotes a uniform distribution in the range \([a, b]\).
    \item The scaling factors \( \alpha = 0.5 \) and \( \beta = 0.2 \) balance depth variation effectively.
\end{itemize}

\noindent \textbf{Dataset-Specific Adaptation:}  
For UTDAC2020, we employed an adaptive Depth Jitter strategy based on depth variance, selectively augmenting only high-variance images while leaving low-variance ones unaltered. This method ensured meaningful augmentation without introducing redundant transformations.

For FathomNet, a fixed depth offset was applied across all images, as preliminary experiments revealed that a random uniform offset between \([-4m, 15m]\) yielded the most stable improvements. Unlike UTDAC2020, where depth variance determined augmentation, this approach allowed a controlled introduction of depth variation in a dataset where more consistent augmentation performed better.

\section{Results and Discussion}

\subsection{Experimental Setup}
\noindent To evaluate the effectiveness of the proposed Depth Jitter augmentation, we conducted multi-label classification experiments using the Query2Label (Q2L)~\cite{query2label} transformer-based framework. All experiments were implemented in PyTorch Lightning and executed on two NVIDIA A40 GPUs, each with 48GB of memory. The model architecture employed a ResNeSt-101 backbone, with input images resized to $512 \times 512$ for the UTDAC2020 dataset and $384 \times 384$ for FathomNet to accommodate their respective resolutions and content densities.

\noindent We adopted the AdamW optimizer with a OneCycleLR learning rate scheduler to ensure stable convergence, and used a batch size of 128 to balance efficiency and memory usage. Asymmetric Loss (ASL)~\cite{ben2020asymmetric} was selected as the objective function due to its effectiveness in handling label imbalance, which is common in underwater multi-label datasets. We compared Depth Jitter against several commonly used augmentation strategies, including a baseline with no augmentation, ColorJitter, CLAHE (Contrast Limited Adaptive Histogram Equalization), and a combined variant of Depth Jitter with ColorJitter. Evaluation was based on multiple performance metrics, including mean Average Precision (mAP), mAP@20, Receiver Operating Characteristic Area Under the Curve (ROC AUC), Precision, Recall, F1 Score, and Validation Loss.

\subsection{Benchmark on UTDAC2020}
\noindent On UTDAC2020, we employ an adaptive Depth Jitter strategy using a quantile-based threshold to identify images with high depth variance. These images receive dynamic depth offsets, simulating realistic underwater conditions. As per the dataset authors, mean Average Precision (mAP) serves as the primary evaluation metric.

\begin{table}[!th]
\centering
\small
\resizebox{\columnwidth}{!}{%
\begin{tabular}{|l|c|c|c|c|c|c|c|}
\hline
\textbf{Augmentation} & \textbf{mAP ↑} & \textbf{mAP@20 ↑} & \textbf{ROC AUC ↑} & \textbf{Prec ↑} & \textbf{F1 ↑} & \textbf{Rec ↑} & \textbf{Loss ↓} \\
\hline
Baseline & 0.95 & 0.87 & 0.98 & 0.86 & 0.91 & 0.97 & 0.045 \\
ColorJitter & 0.93 & 0.86 & 0.98 & 0.85 & 0.91 & 0.96 & 0.043 \\
CLAHE & 0.96 & 0.85 & 0.97 & 0.85 & 0.90 & 0.97 & 0.042 \\
DJ + CJ & 0.96 & 0.87 & 0.99 & 0.85 & 0.90 & 0.97 & 0.044 \\
\textbf{Depth Jitter} & \textbf{0.97} & \textbf{0.87} & \textbf{0.99} & \textbf{0.87} & \textbf{0.92} & \textbf{0.98} & \textbf{0.042} \\
\hline
\end{tabular}%
}
\caption{Performance on UTDAC2020 using Query2Label with different augmentation strategies.}
\label{tab:utdac_results}
\end{table}

\noindent Depth Jitter achieves the highest mAP (0.97) and ROC AUC (0.99), while also attaining the lowest validation loss (0.042), indicating superior performance and efficient learning. These results highlight the advantage of using physically meaningful, depth-aware perturbations over purely pixel-level augmentations for underwater imagery.

\subsection{Benchmark on FathomNet}
\noindent For FathomNet, we apply a fixed random depth offset in the range $[-4\,\text{m}, 15\,\text{m}]$, identified as optimal through preliminary tuning. Consistent with the dataset’s official evaluation protocol, mAP@20 is used as the main metric.

\begin{table}[!ht]
\centering
\small
\resizebox{\columnwidth}{!}{%
\begin{tabular}{|l|c|c|c|c|c|c|c|}
\hline
\textbf{Augmentation} & \textbf{mAP ↑} & \textbf{mAP@20 ↑} & \textbf{ROC AUC ↑} & \textbf{Prec ↑} & \textbf{F1 ↑} & \textbf{Rec ↑} & \textbf{Loss ↓} \\
\hline
Baseline & 0.80 & 0.85 & 0.97 & 0.72 & 0.74 & 0.78 & 0.16 \\
ColorJitter & 0.81 & 0.86 & 0.92 & \textbf{0.75} & 0.73 & 0.78 & 0.17 \\
CLAHE & 0.80 & 0.86 & 0.98 & 0.71 & 0.74 & 0.77 & 0.18 \\
DJ + CJ & 0.80 & 0.86 & 0.98 & 0.71 & \textbf{0.75} & 0.78 & 0.17 \\
\textbf{Depth Jitter} & \textbf{0.81} & \textbf{0.87} & \textbf{0.99} & 0.70 & 0.74 & \textbf{0.79} & \textbf{0.16} \\
\hline
\end{tabular}%
}
\caption{Performance on FathomNet using Query2Label with different augmentation strategies.}
\label{tab:fathomnet_results}
\end{table}

\noindent Depth Jitter again delivers the highest mAP@20 (0.87) and ROC AUC (0.99), along with the best recall (0.79) and lowest loss (0.16). While performance gains over baseline are more modest than in UTDAC2020, the results confirm the method’s robustness across varied underwater contexts and dataset scales.\\
\begin{figure}[!ht]
    \centering
    \includegraphics[width=\linewidth]{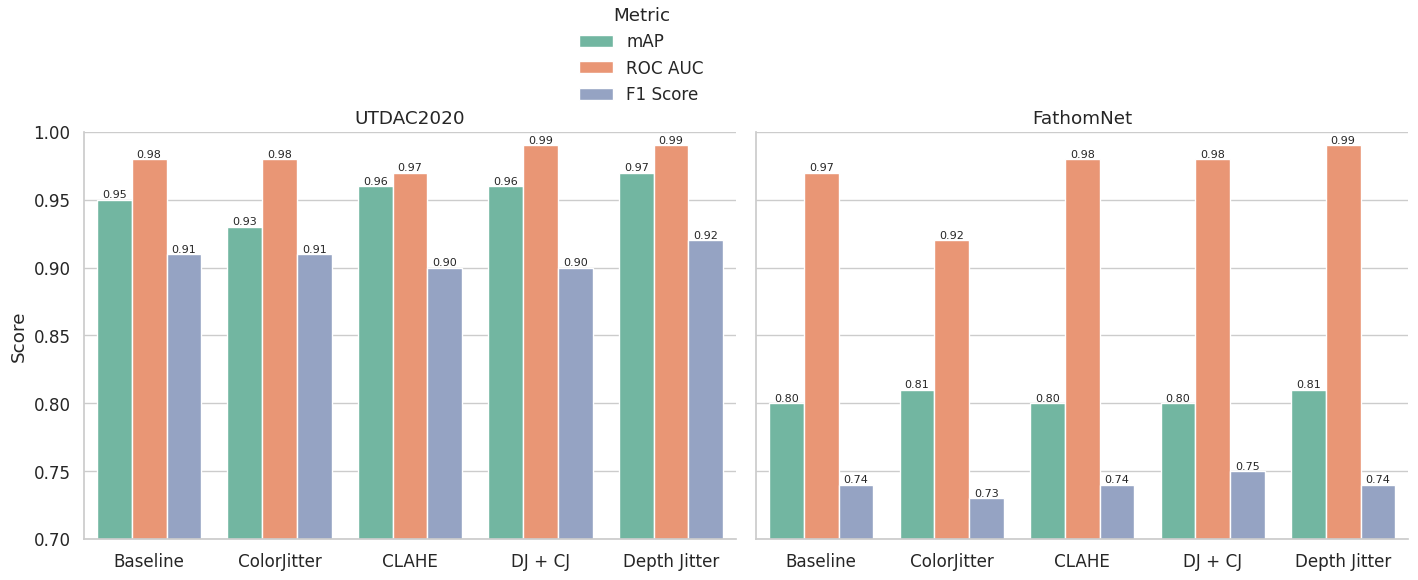}
    \caption{Comparison of mAP, ROC AUC, and F1 Score across augmentation strategies on UTDAC2020 and FathomNet. Depth Jitter (ours) consistently outperforms other methods across all three key metrics.}
    \label{fig:metric_bar_comparison}
\end{figure}
Figure~\ref{fig:metric_bar_comparison} summarizes key performance metrics across datasets and augmentation strategies. Depth Jitter consistently outperforms baselines and enhancement-based methods, particularly in mAP and ROC AUC, confirming its depth-aware advantage.
\subsection{Qualitative Result}
\begin{figure*}[htbp]
    \centering
    \includegraphics[width=\textwidth]{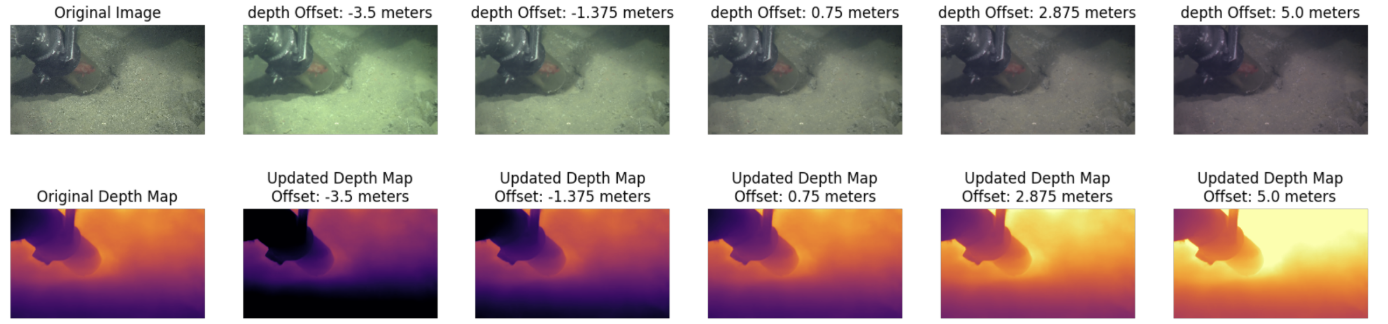}
    \includegraphics[width=\textwidth]{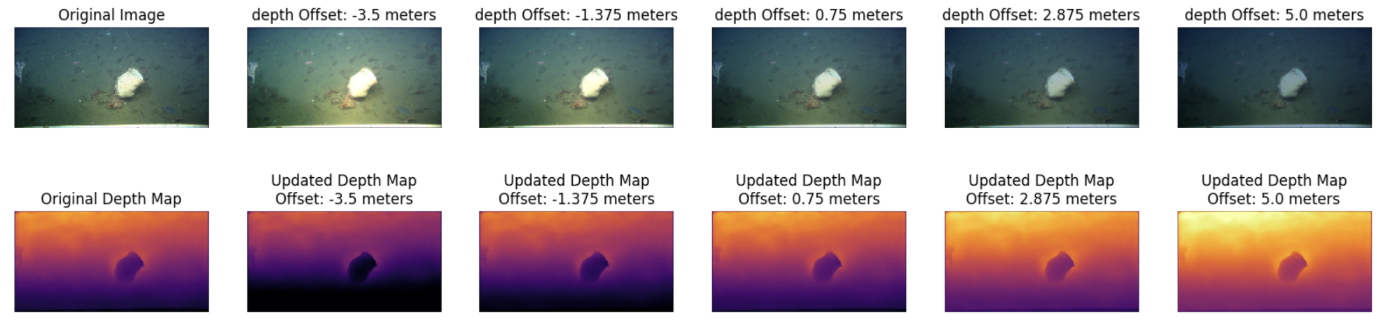}
    \includegraphics[width=\textwidth]{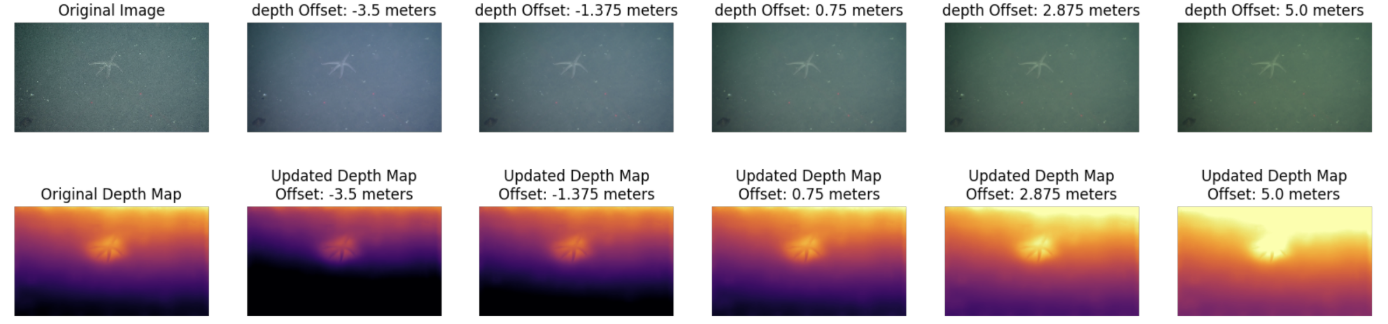}
    \caption{Visualization of Depth Jitter augmentation applied to underwater images. Each row pair shows original RGB images and corresponding depth maps (left), followed by augmented samples with increasing depth offsets ranging from $-3.5$ to $+5.0$ meters. As depth increases, images appear darker and more attenuated, simulating realistic underwater lighting conditions. The updated depth maps reflect the shift in depth distributions used during augmentation. This demonstrates how Depth Jitter introduces structured depth-aware variability while preserving semantic content.}
    \label{fig:depth_jitter_visualization}
\end{figure*}
\noindent Figure~\ref{fig:depth_jitter_visualization} illustrates the impact of Depth Jitter augmentation on underwater images and their corresponding depth maps. Each row represents a sequence of transformations applied to an original image, where different depth offsets are introduced to simulate variations in scene depth.

\noindent The first column in each row displays the original RGB image and its corresponding depth map. Subsequent columns show the results of applying depth offsets ranging from negative to positive values. Negative depth offsets simulate scenarios where the object appears closer, enhancing brightness and contrast due to reduced light absorption and scattering. Conversely, positive depth offsets simulate deeper environments, introducing darker tones and stronger color attenuation, mimicking real-world underwater lighting effects.

\noindent The depth maps in the second row reveal the structural changes resulting from these offsets. As the depth offset increases, the depth estimation progressively shifts, altering pixel intensity distributions. This augmentation strategy enhances model robustness by exposing it to a range of depth conditions, improving generalization in depth-sensitive tasks.

\noindent Overall, Depth Jitter provides a realistic depth-aware augmentation method that enhances training datasets by incorporating natural depth-based variability, making it particularly effective for underwater multi-label classification.
\subsection{Discussion}
\noindent The empirical results across both UTDAC2020 and FathomNet strongly support the effectiveness of Depth Jitter as a depth-aware data augmentation strategy for underwater multi-label classification. On both benchmarks, Depth Jitter consistently outperforms baseline and traditional augmentation methods, achieving higher mAP, mAP@20, ROC AUC, and F1 scores. These improvements validate the central hypothesis of this work: that introducing physically meaningful variations in depth can more effectively capture the complexity of underwater imaging conditions than purely appearance-based perturbations.

\noindent Unlike augmentations such as ColorJitter or CLAHE, which manipulate pixel values without regard to the scene’s physical structure, Depth Jitter simulates realistic changes in scene depth by shifting estimated depth maps within a plausible range. This process mimics the natural effects of light absorption, scattering, and color attenuation in underwater environments. As a result, the model is exposed to a broader distribution of lighting conditions and contrast variations that are grounded in real-world physics, enhancing its generalization ability.

\noindent One of the noteworthy strengths of Depth Jitter lies in its adaptability. The method was tuned differently for each dataset to best match the inherent variability and annotation characteristics. On UTDAC2020, where scene complexity and depth variation are high, an adaptive augmentation strategy was employed using quantile-based thresholds to identify high-variance regions. For FathomNet, which contains a much larger volume of ecologically annotated data collected by ROVs, a simpler fixed random offset in the range $[-4\,\text{m}, 15\,\text{m}]$ yielded optimal results. This flexibility indicates that Depth Jitter can be customized to different imaging setups and sensor conditions, making it suitable for a wide range of marine vision tasks.

\noindent In terms of computational cost, Depth Jitter introduces a moderate increase in training time, typically adding 40 to 60 seconds per epoch. However, this additional overhead is relatively minor compared to the performance gains achieved and is further mitigated by the use of precomputed depth offset parameters. The augmentation process itself is highly efficient during runtime, requiring no additional model components or architectural changes. This makes it a practical solution for researchers and practitioners looking to enhance underwater datasets without significant infrastructure modifications.

\noindent Overall, Depth Jitter bridges the gap between synthetic augmentation and ecological realism. By incorporating domain knowledge from the physics of underwater image formation, it produces augmented samples that better reflect the true visual variability encountered in marine environments. These results not only underscore the value of depth-aware augmentation but also highlight the importance of designing task-specific augmentation strategies grounded in physical context, particularly for specialized domains like underwater vision.

\noindent Overall, Depth Jitter effectively augments underwater images by simulating depth-dependent variations, improving classification robustness in challenging visual conditions.

\subsection{Limitations}
\noindent While Depth Jitter demonstrates significant improvements across multiple evaluation metrics, several limitations remain that offer opportunities for future exploration.

\noindent First, the augmentation strategy relies on the availability and quality of depth maps. In this study, we use estimated depth maps from a pre-trained monocular depth model, which may introduce inaccuracies in regions with complex geometry or poor visibility. Consequently, the effectiveness of Depth Jitter is partially dependent on the fidelity of these depth estimates, especially in turbid or low-light underwater conditions where depth prediction models may struggle.

\noindent Second, the current implementation applies global or region-specific depth perturbations based on statistical heuristics (e.g., quantiles). Although effective, these perturbations are not learned and may not capture task-specific contextual nuances. Incorporating a learnable offset generation mechanism, possibly guided by uncertainty or class distribution, could further enhance performance.

\noindent Third, Depth Jitter introduces moderate computational overhead, primarily during the preprocessing stage when depth offsets are applied to augment training data. Although this cost is amortized over training and kept minimal via precomputation, real-time or embedded systems may still find it constraining without further optimization.

\noindent Finally, our evaluation is limited to two underwater image classification datasets. While they offer diversity in content and acquisition settings, further validation across broader domains (e.g., sonar, medical, or aerial depth-sensitive imagery) would help assess the generalizability of the method.

\section{Conclusion \& Future Work}
\noindent This study introduced Depth Jitter, a novel depth-aware augmentation technique for improving underwater multi-label classification using Query2Label. By simulating realistic depth-dependent variations, Depth Jitter effectively mitigated challenges associated with poor visibility, lighting inconsistencies, and color distortions in underwater imagery. The results demonstrated that Depth Jitter consistently outperformed conventional augmentation strategies, achieving the highest mAP@20 on FathomNet and mAP on UTDAC2020, validating its effectiveness for real-world marine research applications.

\noindent Despite its advantages, Depth Jitter has some limitations. The augmentation relies on estimated depth maps, which can introduce errors due to inaccuracies in depth inference models. Additionally, the approach assumes a uniform underwater environment, which may not generalize well to highly dynamic marine conditions with varying turbidity levels. Computational overhead is another concern, as precomputing underwater image formation parameters is required for efficient augmentation.

\noindent Future research should address these limitations by exploring adaptive depth estimation techniques that dynamically adjust to environmental variations. Self-supervised learning or diffusion-based augmentation could further enhance model robustness by learning better depth representations. Additionally, expanding datasets with geographically diverse underwater imagery and integrating domain adaptation techniques will improve model generalization. Finally, advancing multi-label classification methods with graph-based reasoning or attention mechanisms could further refine species recognition, contributing to more accurate biodiversity monitoring and marine conservation efforts.

\section*{Acknowledgment}
This research was funded by the LIS Lab at Université de Toulon and LARSyS FCT funding (DOI: 10.54499/LA/P/0083/2020, 10.54499/UIDP/50009/2020, and 10.54499/UIDB/50009/2020). We extend our gratitude to colleagues, friends, and family for their support throughout this process.
\bibliographystyle{ieeetr}
\bibliography{conference_101719}
\end{document}